\documentclass{article}

\usepackage{arxiv}

\usepackage[utf8]{inputenc} 
\usepackage[T1]{fontenc}    
\usepackage{hyperref}       
\usepackage{url}            
\usepackage{booktabs}       
\usepackage{amsfonts}       
\usepackage{nicefrac}       
\usepackage{microtype}      
\usepackage{lipsum}
\usepackage{svg}
\usepackage{graphicx}
\graphicspath{ {./images/} }
\usepackage{amsmath} 
\usepackage{pdflscape}
\usepackage{xcolor}
\usepackage[normalem]{ulem}
\usepackage{comment}
\usepackage{longtable}
\usepackage{tabularx}
\usepackage{caption} 
\usepackage{subcaption}
\usepackage{longtable}
\usepackage{booktabs}
\usepackage[utf8]{inputenc}
\usepackage{changepage}

\usepackage{array}
\usepackage{float}
\newlength{\extralength}
\newlength{\fulllength}
\newcolumntype{C}{>{\centering\arraybackslash}X}

\title{Fall Detection for Industrial Setups using YOLOv8 variants}

\author{Gracile Astlin Pereira \\[1ex]
\begin{minipage}[t]{0.90\textwidth}
\centering
\scriptsize Department of Computer Science, Huddersfield University, Queensgate, Huddersfield HD1 3DH, UK \\
Correspondence: U2292824@unimail.hud.ac.uk
\end{minipage}}

\begin{document}

\maketitle
\begin{abstract} This paper presents the development of an industrial fall detection system utilizing YOLOv8 variants, enhanced by our proposed augmentation pipeline to increase dataset variance and improve detection accuracy. Among the models evaluated, the YOLOv8m model, consisting of 25.9 million parameters and 79.1 GFLOPs, demonstrated a respectable balance between computational efficiency and detection performance, achieving a mean Average Precision (mAP) of 0.971 at 50\% Intersection over Union (IoU) across both "Fall Detected" and "Human in Motion" categories. Although the YOLOv8l and YOLOv8x models presented higher precision and recall, particularly in fall detection, their higher computational demands and model size make them less suitable for resource-constrained environments.
\end{abstract}


\section{Introduction}

Computer vision is a subset of Artificial Intelligence which enables automated visual intelligence. There are several domains that are looking towards automated visual inspections from warehousing \cite{hussain2023yolo} to defect detection in photovoltaics \cite{hussain2023review}. The benefits of automated visual inspection include reduced labour costs, improved accuracy \cite{aydin2023domain}, less bias \cite{hussain2023child} and reduced hardware costs \cite{alsboui2022dynamic}.
In industrial environments, ensuring the safety of workers is important, particularly in sectors where falls are a significant risk \cite{iparraguirre2023improving}. Traditional fall detection systems \cite{vallejo2013artificial, hsieh2017novel} have relied on various methods like accelerometers and gyroscopes, infrared or pressure pads, including simple threshold-based algorithms \cite{mouratidis2005threshold} and earlier generations of object detection models. However, these approaches often fall short in terms of accuracy, real-time performance, and adaptability to complex industrial settings. Historically, fall detection systems have utilised models such as YOLOv3 \cite{redmon2018yolov3} and YOLOv4 \cite{bochkovskiy2020yolov4}. While these models have provided valuable contributions, they are limited by their outdated architectures and lack of advancements in feature extraction and real-time processing capabilities \cite{arip2024object}. YOLOv3, for example, although revolutionary at its introduction, struggles with finer details and contextual understanding in dynamic environments. YOLOv4 improved upon this by offering better accuracy and efficiency but still faced challenges with the evolving demands of modern industrial settings.

Following these, YOLOv5 introduced a series of improvements, including a more refined architecture and enhanced flexibility in training and deployment. YOLOv5's modular design allowed for easier customisation and fine-tuning, making it well-suited for various applications, including fall detection. It offered improved accuracy and speed compared to YOLOv4, but still faced challenges in very dynamic or cluttered environments. 
YOLOv6 \cite{li2022yolov6} continued the evolution with further optimisations, enhancing both detection speed and accuracy. It featured a more efficient model design and better integration of modern techniques such as transformer-based layers. YOLOv6 improved the robustness of object detection, yet it still had limitations in certain complex scenarios and required additional fine-tuning for optimal performance in specific industrial settings. 
YOLOv7 \cite{wang2023yolov7}, the successor in this series, brought significant advances in real-time performance and accuracy. It introduced innovative techniques such as the EfficientDet backbone and the use of attention mechanisms, which enhanced its ability to detect objects with greater precision. However, while YOLOv7 improved on many aspects, it still had room for improvement in handling highly variable and complex industrial environments. 
After the introduction of YOLOv8 \cite{terven2023comprehensive}, it addresses many limitations of its predecessors. It incorporates advanced features such as improved backbone networks for feature extraction, enhanced efficiency in processing, and a more robust architecture for handling diverse and challenging scenarios in object detection \cite{hu2024macnet}. By capturing richer object feature information, YOLOv8 can effectively identify diverse and complex objects in different environments, thereby enhancing detector performance \cite{wang2023detection}. YOLOv8 represents the latest advancement in the YOLO detection network, achieving notable improvements in both detection speed and accuracy \cite{tong2024wtbd}. This iteration of the YOLO model series notably enhances efficiency and performance in real-time object detection.

In summary, while earlier models laid a strong foundation for object detection and fall detection, YOLOv8's advanced features and enhancements make it the premier choice for modern industrial applications. Its state-of-the-art capabilities offer a significant advancement in fall detection technology, ensuring a safer working environment. A detailed review of related works is provided in the following section.

\section{Literature Review} 

Falls in industrial settings, such as manufacturing plants, construction sites, and warehouses, pose significant risks to worker safety. Effective fall detection systems are crucial for mitigating these risks and improving workplace safety. This review explores current technologies and methodologies used for fall detection in industrial environments, examining their effectiveness, challenges, and future directions.

Singh, Rajput and Sharma's \cite{singh2020human} article on human fall detection systems highlights three main approaches: wearable devices, context-aware systems, and vision-based technologies. Wearable systems often face challenges with user compliance and sensor limitations, while context-aware methods struggle with environmental noise and integration complexity. Vision-based systems, utilising cameras and advanced machine learning algorithms such as CNNs, LSTMs, and deep belief networks, offer significant benefits due to their non-intrusiveness and rich data sources. Recent research indicates that vision-based systems achieve high accuracy, with some models reaching up to 99.73\% accuracy for fall detection. These systems, despite concerns about privacy and processing complexity, demonstrate superior performance in identifying falls compared to other methods. Combining vision-based technology with additional sensors and addressing scalability and privacy issues will be crucial for advancing future fall detection solutions.

The CNN-3B3Conv model, designed by Santos et al. \cite{santos2019accelerometer} for fall detection, leverages three convolutional blocks with varying kernel sizes and a final set of fully-connected layers to classify events as "fall" or "not fall." Trained using stochastic gradient descent and data augmentation techniques, the model was rigorously evaluated against LSTM-based approaches and tested on URFD, SmartWatch, and Notch data sets. Results showed that CNN-3B3Conv with data augmentation achieved 99.86\% accuracy and 100\% precision, surpassing the LSTM models, which achieved up to 98.57\% accuracy. The simpler CNN-1Conv model also outperformed CNN-3B3Conv in some metrics, highlighting the impact of model complexity. The study highlights the effectiveness of data augmentation and simpler models while addressing challenges such as data standardisation and real-world performance. 

Shojaei-Hashemi et al. \cite{shojaei2018video} presents a novel approach for human fall detection in smart homes using deep learning, specifically long short-term memory (LSTM) neural networks. Unlike previous methods reliant on manually crafted features, this model automatically extracts relevant features from depth camera data, addressing privacy concerns associated with traditional video methods. By employing transfer learning, the model compensates for the limited availability of fall samples by pre-training on a larger dataset of regular actions. Evaluated on the NTU RGB+D Action Recognition Dataset, the proposed method achieved a high performance with an area under the ROC curve (AUC) of 0.99, and demonstrated superior results with 93\% precision and 96\% recall compared to existing methods. This approach highlights the effectiveness of deep learning and transfer learning in enhancing fall detection capabilities in smart home settings.

YOLOv3 \cite{redmon2018yolov3} processes images at multiple scales to generate feature maps used for classification \cite{wang2020human}. It also detects people, with features extracted by a CNN and analysed over time by an LSTM network for event classification \cite{feng2020spatio}. Additionally, YOLO integrates with DeepSORT \cite{wojke2017simple} for tracking multiple objects \cite{wang2020fall}, and it detects fallen individuals by rotating images to maximise sensitivity, comparing bounding boxes for classification \cite{maldonado2019fallen}. These methods showcase YOLO's adaptability in real-time detection, classification, and tracking tasks.

Zheng et al. \cite{zheng2024high} reviewed advancements in object detection models with a specific focus on YOLOv8 enhancements for fall detection. Their work discusses the integration of triplet attention mechanisms, which notably improve model accuracy and robustness, with the yolov8s\_Tri\_Multi configuration achieving the highest accuracy improvements of 0.4\% and 0.6\% for mAP@0.5 and mAP@0.5–0.95, respectively. The incorporation of FasterNet and deformable convolution modules, alongside the triplet attention, further enhances feature extraction and multi-scale fusion, leading to a significant performance boost. Comparative results reveal that FDT-YOLO surpasses other mainstream object detection models, such as Faster R-CNN, SSD, and various YOLO versions, with mAP@0.5 reaching 96.3\% and mAP@0.5–0.95 at 85.9\%, while maintaining a low parameter count of 9.9 million. The model's robust performance under varied conditions, including different light levels and occlusions, highlights its suitability for practical deployment, particularly in mobile and edge devices. Future work aims to validate the model’s scalability and efficiency in real-world elderly care scenarios.

Another work by Saricayir, Alican and Ozcan \cite{sariccayir1deep} presents a deep learning-based fall detection system using YOLOv7 and YOLOv8 architectures to enhance real-time assistance for the elderly. It highlights the growing public health issue of falls among older adults, especially in care settings, and introduces a novel method for detecting and localising falls in video streams. The study employs a dataset with various human activities and evaluates the models using metrics such as mean Average Precision (mAP), finding YOLOv7 superior in accuracy but YOLOv8 faster. Despite YOLOv7's higher accuracy and robustness, the study acknowledges limitations, including dataset quality and the risk of false positives. The proposed system aims to improve safety and quality of life by providing rapid intervention and real-time information to caregivers.

\section{Methodology}
\subsection{Dataset}
\begin{figure}[H]
\begin{adjustwidth}{-\extralength}{0cm}
\centering
\includegraphics[width=15cm]{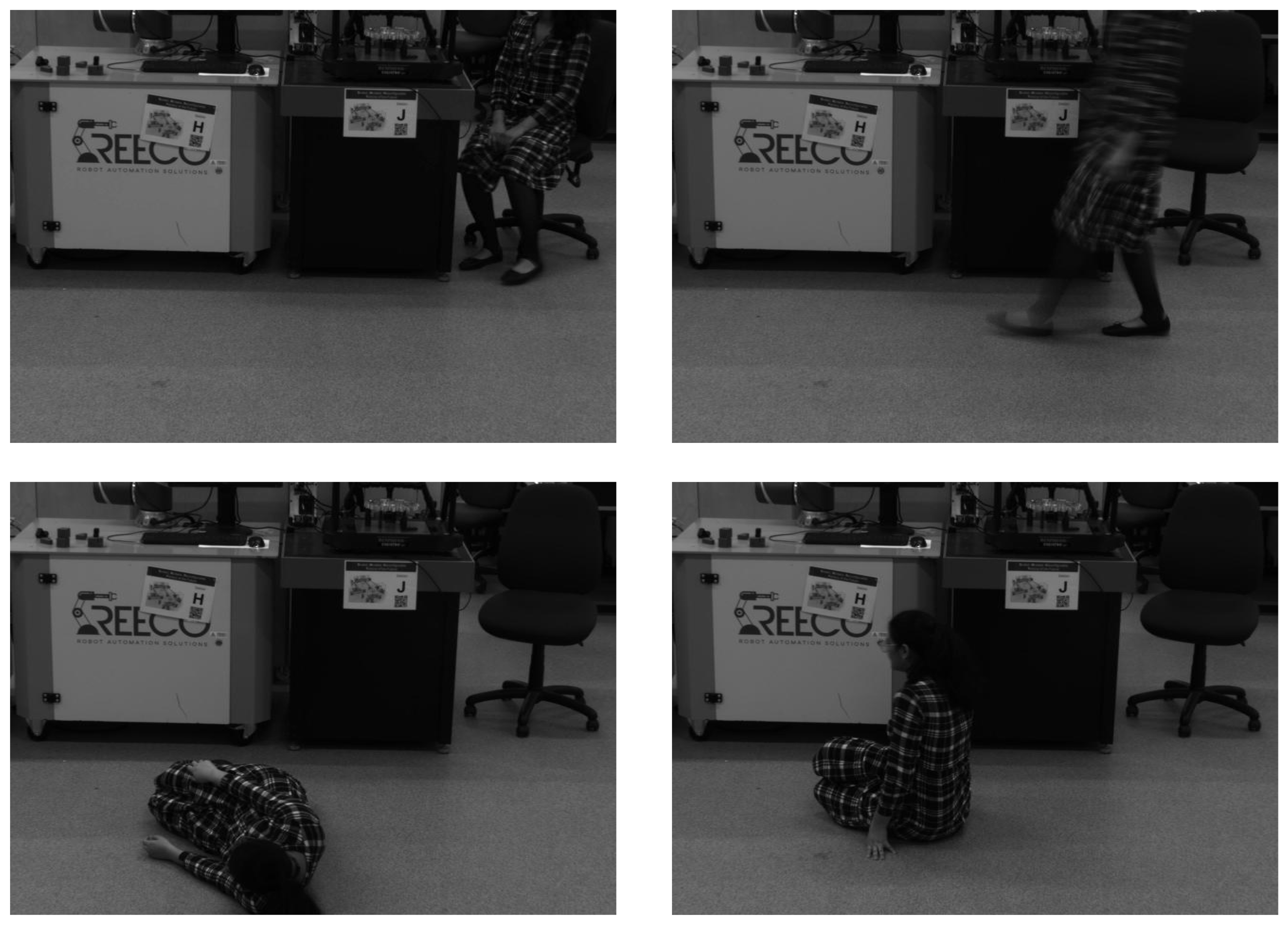}
\end{adjustwidth}
\caption{Industrial Activity Dataset}
\label{Figure:1}
\end{figure}

The dataset was derived from high-resolution video (2560x1920 pixels) recorded at 6 FPS using an 8-bit, 5 MP IDS UI-1485LE-M-GL monochrome camera, mounted on a wall in an industrial setup within the university lab and equipped with a CS mount. The IDS camera mounted on the wall was connected to a Jetson Nano. The Jetson Nano was then linked to a local computer to capture and save the video for further machine learning processing. The saved video was subsequently processed and divided into 95 distinct frames using a python script. Each of the 95 frames extracted from the video were subjected to manual annotation with Label Studio, categorising them into two distinct classes: human in motion, and fall detected. A selection of frames from the dataset is illustrated in figure \ref{Figure:1}.

\subsection{Data Augmentation}
Data augmentation is an essential preprocessing method in machine learning and computer vision that enhances the accuracy and durability of models. To overcome the constraints of a small dataset, various augmentation strategies were applied to generate additional images, thereby broadening the range of training data and reducing overfitting by preventing the model from memorising specific details. Furthermore, augmentation addresses class imbalance by increasing the number of images in less-represented categories, which improves the overall classification performance and reduces biases. Due to the video being recorded at a low frame rate of 6 FPS, the process of extracting individual frames revealed significant challenges, notably the presence of blurriness in many frames. This blurriness is attributable to the insufficient capture speed, which leads to motion artifacts and a lack of temporal resolution. Consequently, the quality and clarity of the extracted frames are compromised, impacting the accuracy and reliability of subsequent analyses. To improve model performance with the available dataset, various augmentation techniques were employed. Basic transformations like auto-orientation, random resizing, and gray-scaling were executed using Roboflow. Furthermore, additional augmentations, including blur, median blur, and CLAHE (Contrast Limited Adaptive Histogram Equalisation), were implemented during the model training phase in Ultralytics, utilising the Albumentations library.

\subsubsection{Random Resize}
\begin{figure}[H]
\begin{adjustwidth}{-\extralength}{0cm}
\centering
\includegraphics[width=12cm]{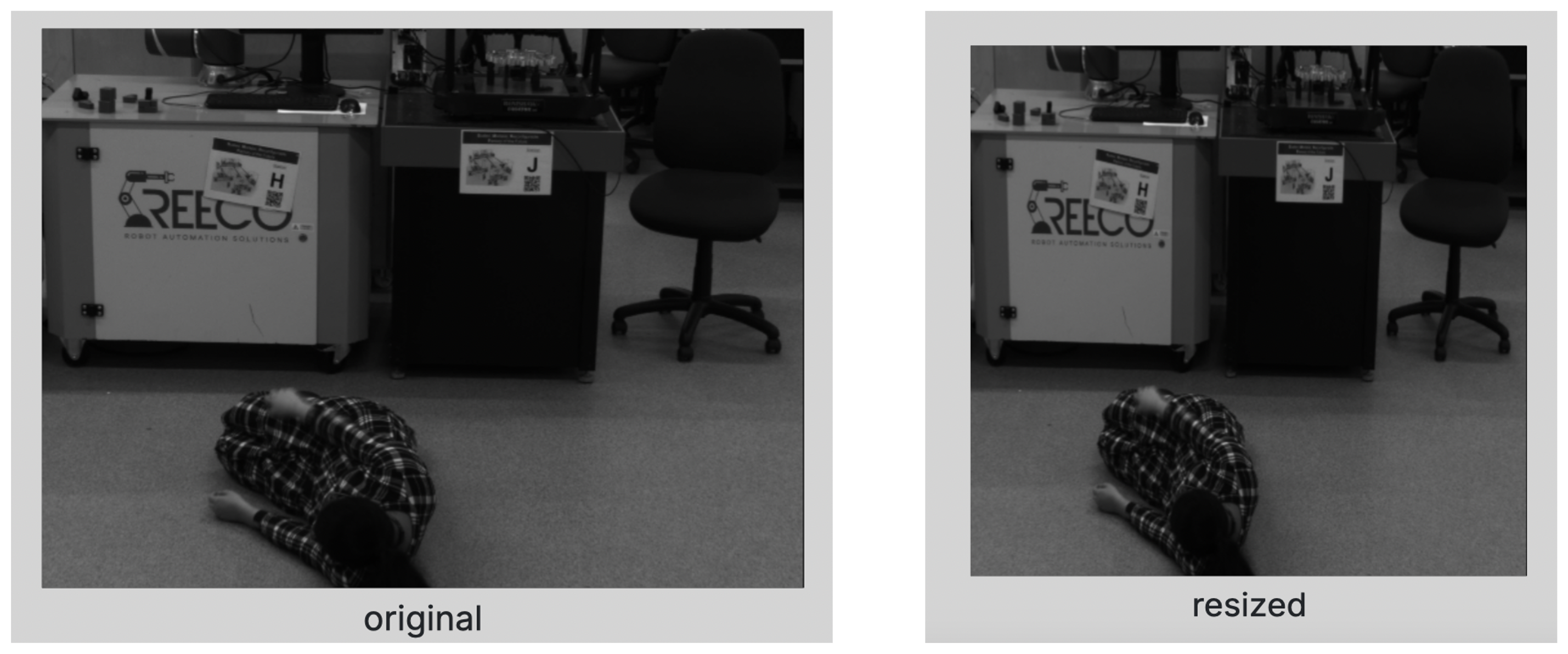}
\end{adjustwidth}
\caption{Image after random resizing}
\label{Figure:2}
\end{figure}

Resizing images is a fundamental technique in data augmentation for machine learning, particularly in training convolutional neural networks (CNNs) for tasks such as image classification, object detection, and segmentation. It plays a crucial role in standardising input dimensions to ensure consistency during model training \cite{ma2014depth}. As demonstrated in figure \ref{Figure:2}, using Roboflow, the images were randomly resized from 2560 x 1920 pixels to 640 x 640 pixels to maintain consistent dimensions and ensure model consistency. By resizing images, the model can better generalise and recognise objects at various scales, thereby enhancing performance on unseen data. Additionally, resizing contributes to enhancing the diversity of the training dataset, making the model more robust to real-world variations. Several common methods for resizing images include rescaling by a constant factor using interpolation techniques like bilinear or bicubic interpolation, as well as random resizing to introduce variability and enhance the model's adaptability to different object sizes and perspectives \cite{salimi2022using}. 

\subsubsection{Random Grayscale}
\begin{figure}[H]
\begin{adjustwidth}{-\extralength}{0cm}
\centering
\includegraphics[width=12cm]{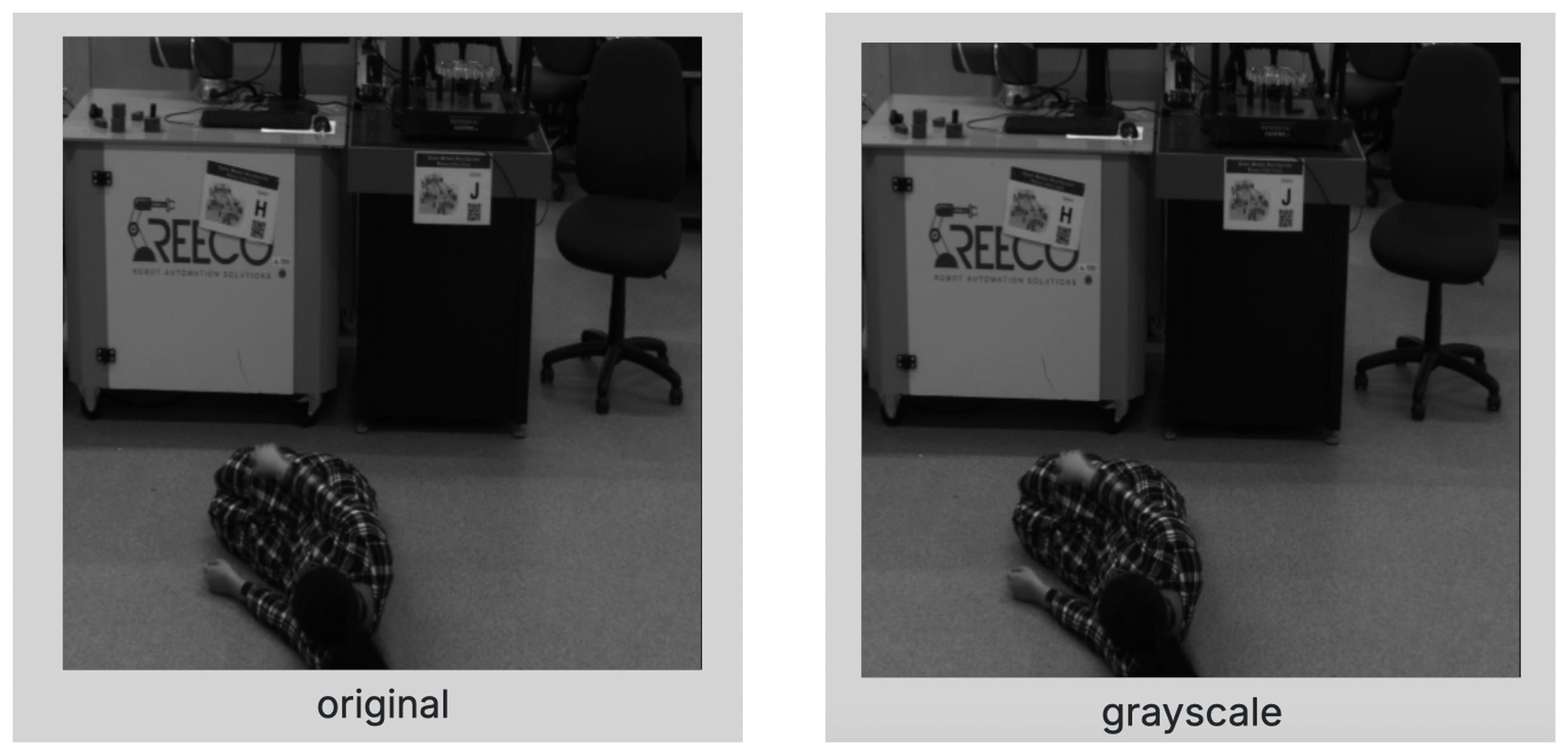}
\end{adjustwidth}
\caption{Image after adding gray scaling}
\label{Figure:3}
\end{figure}

The ToGray augmentation is applied with a low probability of p=0.01, meaning only 15\% of the images will be converted to grayscale during training. By training on grayscale images using Roboflow, the model learns to identify objects and patterns based solely on intensity rather than color. This can enhance the model's robustness to variations in color and lighting conditions, as it becomes more focused on shape and texture features that are independent of color, as displayed in figure \ref{Figure:3}. By converting the images to grayscale, the model's ability to identify and assess hazards is improved, making it more effective in diverse environments where color information is not available.

\subsubsection{Blur}
\begin{figure}[H]
\begin{adjustwidth}{-\extralength}{0cm}
\centering
\includegraphics[width=12cm]{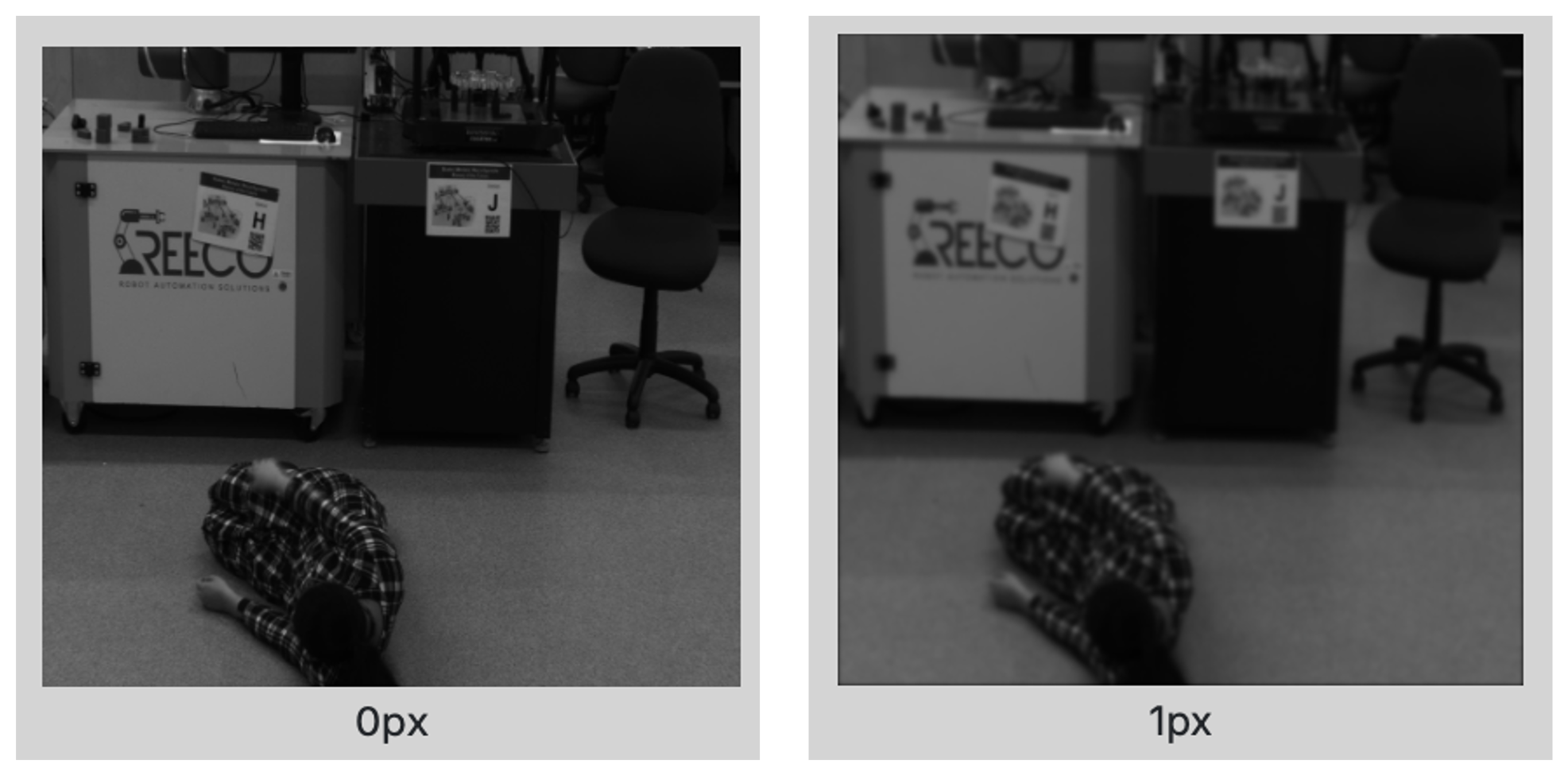}
\end{adjustwidth}
\caption{Blur effect}
\label{Figure:4}
\end{figure}

The Blur augmentation applies a blur effect to the image, which softens its details and reduces noise. This strategy is controlled by a probability parameter, $p=0.01$, meaning that the blur effect is applied to only 1\% of the images during training. The intensity of the blur is determined by the \texttt{blur\_limit} parameter, which specifies a range for the blur kernel size. In this case as shown in figure \ref{Figure:4}, the kernel size is chosen randomly between 3 and 7 pixels. By incorporating blur into the training process using albumentations library, the model learns to recognise objects even when they are not perfectly sharp, helping it to generalise better to practical scenarios where images may not always be perfectly clear.

\subsubsection{Median Blur}
\begin{figure}[H]
\begin{adjustwidth}{-\extralength}{0cm}
\centering
\includegraphics[width=12cm]{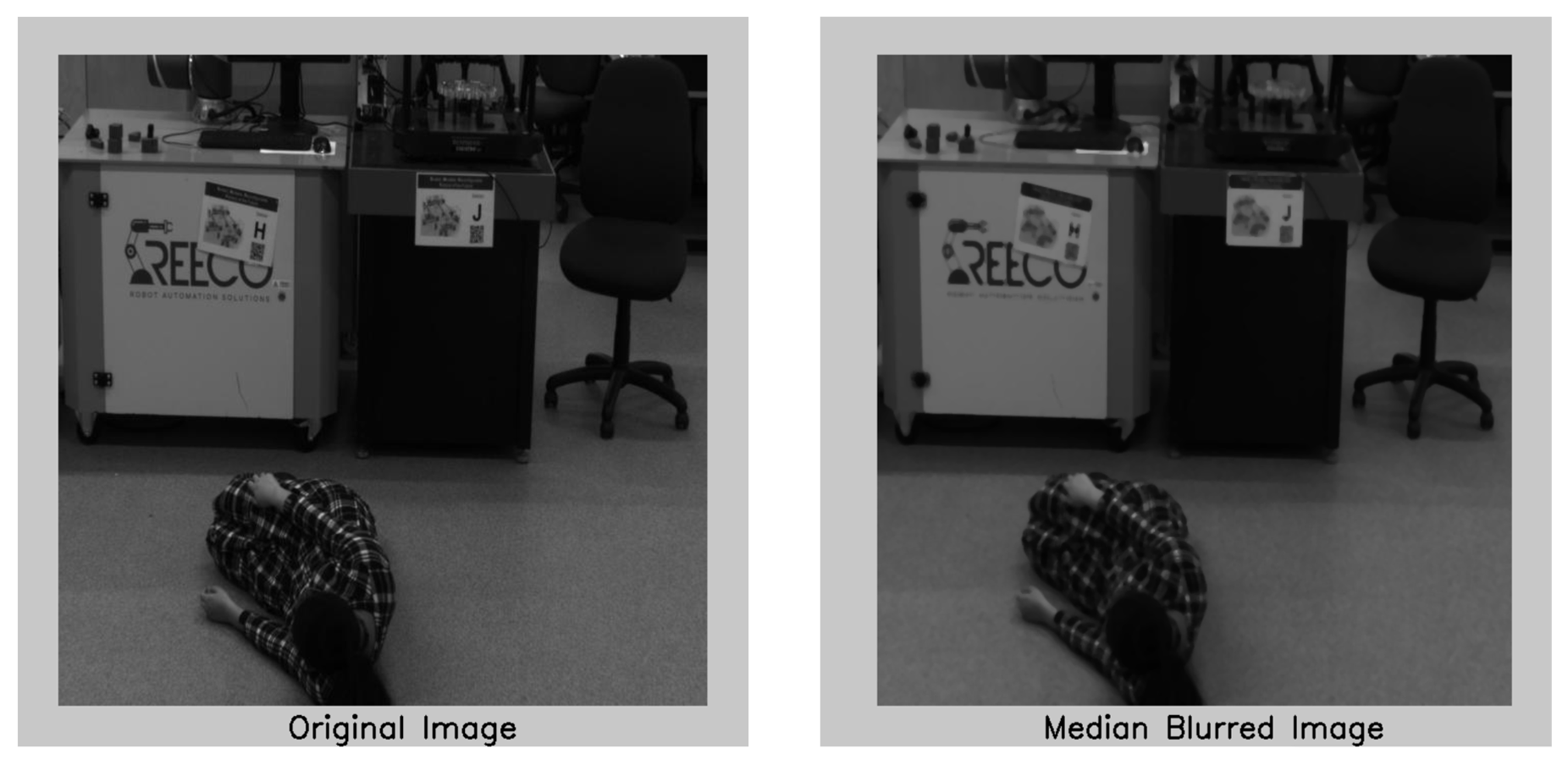}
\end{adjustwidth}
\caption{Median Blur effect}
\label{Figure:5}
\end{figure}

MedianBlur is another image transformation that helps to reduce noise and smooth out irregularities in images. This method replaces each pixel in the image with the median value of the pixels in its local neighborhood. Similar to blur, median blur is applied with a probability of $p=0.01$, indicating that it will affect only 1\% of the training images, this is illustrated in figure \ref{Figure:5}. The extent of the blur is controlled by the \texttt{blur\_limit} parameter, which defines the size of the neighborhood used for calculating the median. The neighborhood size is randomly chosen between 3 and 7 pixels. Median blur is particularly effective in preserving edges while removing salt-and-pepper noise, which can make the model more robust to noisy data.

\subsubsection{CLAHE (Contrast Limited Adaptive Histogram Equalisation)}
\begin{figure}[H]
\begin{adjustwidth}{-\extralength}{0cm}
\centering
\includegraphics[width=12cm]{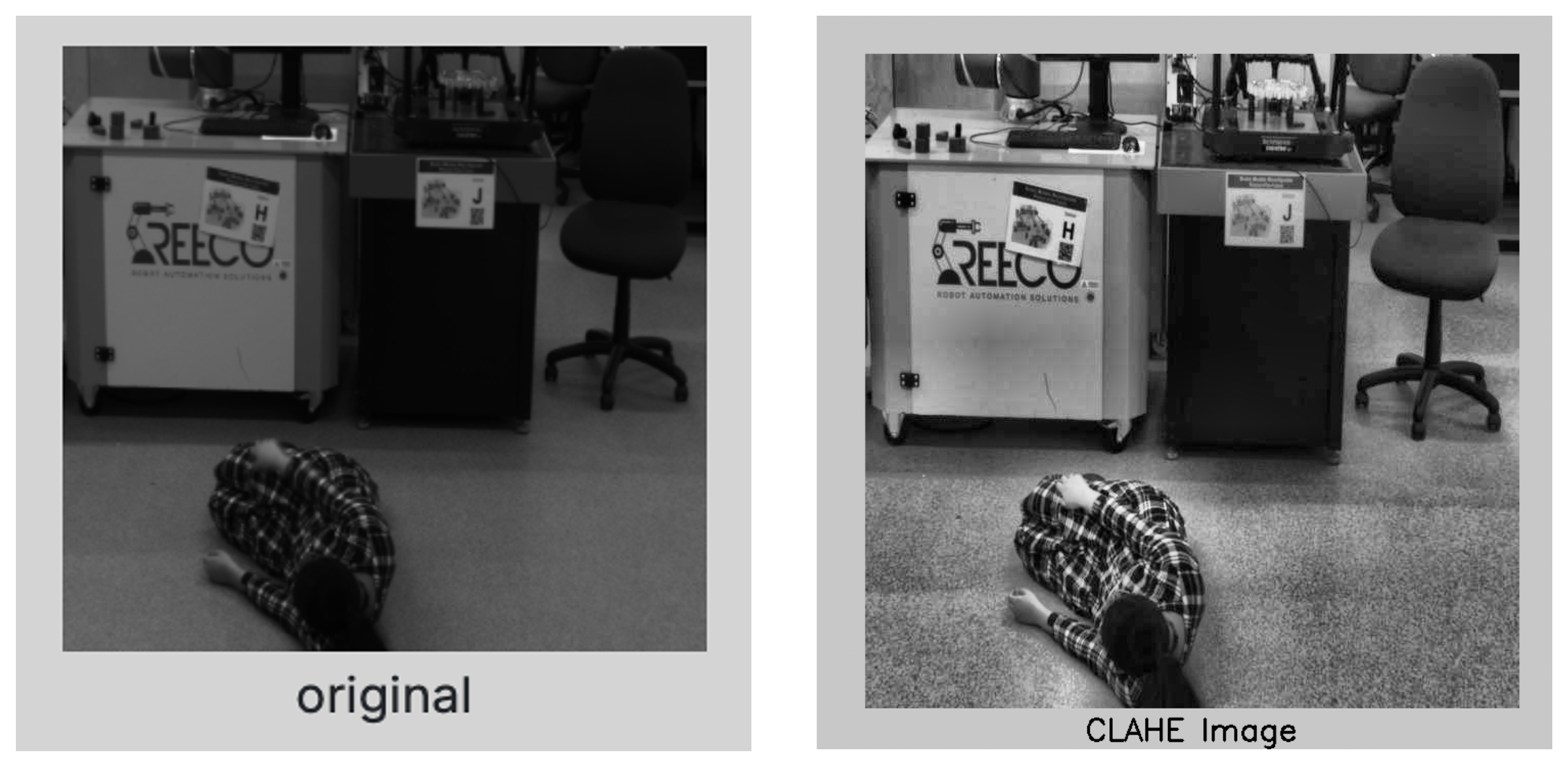}
\end{adjustwidth}
\caption{CLAHE effect}
\label{Figure:6}
\end{figure}

CLAHE is a technique for enhancing image contrast by performing histogram equalisation within small localised regions of the image.
This approach divides the image into tiles and adjusts the contrast locally within each tile. The probability of applying CLAHE is $p=0.01$, meaning that it will be used in 1\% of the images. The \texttt{clip\_limit} parameter sets a threshold for contrast enhancement, with a value randomly chosen between 1 and 4.0, preventing excessive contrast adjustments that might lead to artifacts. The \texttt{tile\_grid\_size} parameter, set to (8, 8), specifies the size of the grid tiles. This is demonstrated in figure \ref{Figure:6}. CLAHE helps improve the visibility of features in images with varying lighting conditions, making the model more adept at handling diverse image qualities. 

Data augmentation techniques like random resize, grayscale, blur, median blur, and CLAHE enhance dataset by introducing variability. Since part of the augmentation was manually performed using Roboflow, the total number of images, after resizing and applying grayscale, amounted to 115. The remaining transformations were applied directly to the training dataset using albumentations.

\subsection{YOLOv8 Architecture}
YOLOv8 (You Only Look Once version 8) improves upon YOLOv5 through several architectural enhancements that elevate detection accuracy. The model integrates the C2f module, which supports an anchorless structure. This change simplifies bounding box predictions and reduces dependency on non-maximum suppression by eliminating traditional anchor boxes that often misrepresent patterns within the dataset \cite{terven2023comprehensive}. The architectural table \ref{tab:Table 1} for our best performing model describes the specifics of each layer’s setup, showcasing how the C2f module and Spatial Pyramid Pooling Fractional (SPPF) refine both location accuracy and feature representation, while also detailing channel dimensions, sizes, and parameters.

The model begins with a series of convolutional layers, such as Conv1 and Conv2, that progressively increase the depth of feature maps while reducing spatial resolution. These initial layers capture basic features from the input image, which are further refined by deeper convolutional layers (Conv3 to Conv5) and enhanced by C2f modules (C2f1 - C2f4) that improve feature representation without altering channel dimensions.
To capture and integrate features at different scales, YOLOv8 uses upsampling (Upsample1 and Upsample2) and concatenation (Concat1 - Concat4) strategies. After applying an SPPF (Spatial Pyramid Pooling - Fast) layer for multi-scale context, the network upsamples and merges features through layers like Upsample1 and Concat1, refining them with additional C2f modules (C2f5-C2f8). The final stages include additional convolutions and concatenations, leading to the detection layer (Detect) which utilises anchors to predict object locations and classes. 

\begin{table}[H]
\renewcommand{\arraystretch}{1.3}
\setlength{\tabcolsep}{15pt} 
\caption{YOLOv8n Architecture}
\label{tab:Table 1}
\centering
\begin{tabular}{lcccr}
\hline
\textbf{Layer} & \textbf{Input Channels} & \textbf{Output Channels} & \textbf{Size} & \textbf{Parameters} \\
\hline
Conv1 & 3 & 16 & 3 × 3, stride 2 & 464 \\
Conv2 & 16 & 32 & 3 × 3, stride 2 & 4672 \\
C2f1 & 32 & 32 & - & 7360 \\
Conv3 & 32 & 64 & 3 × 3, stride 2 & 18560 \\
C2f2 & 64 & 64 & - & 49664 \\
Conv4 & 64 & 128 & 3 × 3, stride 2 & 73984 \\
C2f3 & 128 & 128 & - & 197632 \\
Conv5 & 128 & 256 & 3 × 3, stride 2 & 295424 \\
C2f4 & 256 & 256 & - & 460288 \\
SPPF & 256 & 256 & 5 × 5 pooling & 164608 \\
Upsample1 & - & - & Upsample (2 × 2) & 0 \\
Concat1 & - & - & Concat & 0 \\
C2f5 & 384 & 128 & - & 148224 \\
Upsample2 & - & - & Upsample (2 × 2) & 0 \\
Concat2 & - & - & Concat & 0 \\
C2f6 & 192 & 64 & - & 37248 \\
Conv6 & 64 & 64 & 3 × 3, stride 2 & 36992 \\
Concat3 & - & - & Concat & 0 \\
C2f7 & 192 & 128 & - & 123648 \\
Conv7 & 128 & 128 & 3 × 3, stride 2 & 147712 \\
Concat4 & - & - & Concat & 0 \\
C2f8 & 384 & 256 & - & 493056 \\
Detect & - & - & Detect (anchors) & 751702 \\
\hline
\end{tabular}
\end{table}

YOLOv8 makes use of softmax functions to classify an object into its appropriate class and sigmoid functions to determine whether an object is present inside a bounding box. In contrast to earlier YOLO revisions the model enhancements improve the accuracy of bounding box prediction and address the detection of smaller models by introducing Distribution Focused Loss (DFL) \cite{li2020generalized} functions and Complete Intersection over Union (CIoU) \cite{zheng2020distance}. To improve the classification of several classes, classification loss is calculated using binary cross-entropy.
There are five variants of YOLOv8, which were released in January 2023 by Ultralytics: YOLOv8n (nano), YOLOv8s (small), YOLOv8m (medium), YOLOv8l (large), and YOLOv8x (extra big). It can be used for a number of vision tasks, such as tracking, semantic segmentation, posture estimation, object detection, and classification.
The following table \ref{table:yolov8_comparison} presents a comparative overview of the different YOLOv8 model variants: YOLOv8n, YOLOv8s, YOLOv8m, YOLOv8l, and YOLOv8x, highlighting key attributes that influence their performance and complexity. This comparison includes a range of metrics and design specifications: the number of parameters, GFLOPs (giga floating-point operations per second), number of layers, initial and maximum filter sizes, configuration of C2f blocks etc.

\renewcommand{\arraystretch}{1.2} 

\begin{table}[H]
\caption{Comparison of YOLOv8 Models}
\centering
\begin{tabularx}{\linewidth}{X X X X X X}
\hline
\textbf{Attribute} & \textbf{YOLOv8n} & \textbf{YOLOv8s} & \textbf{YOLOv8m} & \textbf{YOLOv8l} & \textbf{YOLOv8x} \\
\hline
Parameters & 3,011,238 & 11,136,374 & 25,857,478 & 43,631,382 & 68,154,534 \\
GFLOPs & 8.2 & 28.6 & 79.1 & 165.4 & 258.1 \\
Layers & 225 & 225 & 295 & 365 & 365 \\
Initial Conv & 16 filters & 32 filters & 48 filters & 64 filters & 80 filters \\
Max Filters & 256 & 512 & 576 & 512 & 640 \\
C2f Blocks & 32, 64, 128, 256 & 64, 128, 256, 512 & 96, 192, 384, 576 & 128, 256, 512, 1024 & 160, 320, 640, 1280 \\
SPPF Filters & 256 & 512 & 576 & 512 & 640 \\
Upsampling Layers & 2 & 2 & 2 & 2 & 2 \\
Concat Layers & 3 & 4 & 5 & 6 & 6 \\
Detection Anchors & [64, 128, 256] & [128, 256, 512] & [192, 384, 576] & [256, 512, 512] & [320, 640, 640] \\
Detection Head & 2 classes, [64, 128, 256] & 2 classes, [128, 256, 512] & 2 classes, [192, 384, 576] & 2 classes, [256, 512, 512] & 2 classes, [320, 640, 640] \\
Detection Layers & 1 & 1 & 1 & 1 & 1 \\
Overall Complexity & Low & Medium & High & Very High & Highest \\
\hline
\end{tabularx}
\label{table:yolov8_comparison}
\end{table}

\section{Results}
This section exclusively discusses the performance of all the variants of YOLOv8 models during training and validation. The YOLOv8.2 models were trained over 100 epochs for each variant on a dataset aimed at detecting "Fall Detected" and "Human in Motion" classes. The training was conducted in Google Colaboratory using NVDIA Tesla T4 GPUs with 12.7 GB of memory, and the training process required around 2 hours to complete for all the variants. The YOLOv8 model was trained using the AdamW optimizer and a well-structured set of hyperparameters, which is known for its effectiveness in handling weight decay and improving generalisation. The initial and final learning rates were both set to 0.01, with momentum at 0.937 and a weight decay of 0.0005 to mitigate overfitting. A warmup strategy was implemented, where the learning rate gradually increased over 3.0 epochs, with a warmup momentum of 0.8 and a warmup bias learning rate of 0.1. The loss function was balanced with a box loss weight of 7.5, class loss at 0.5, DFL at 1.5, pose loss at 12.0, and key object loss at 1.0. 

\subsection{YOLOv8n}

\begin{figure}[H]
\begin{adjustwidth}{-\extralength}{0cm}
\centering
\includegraphics[width=\linewidth]{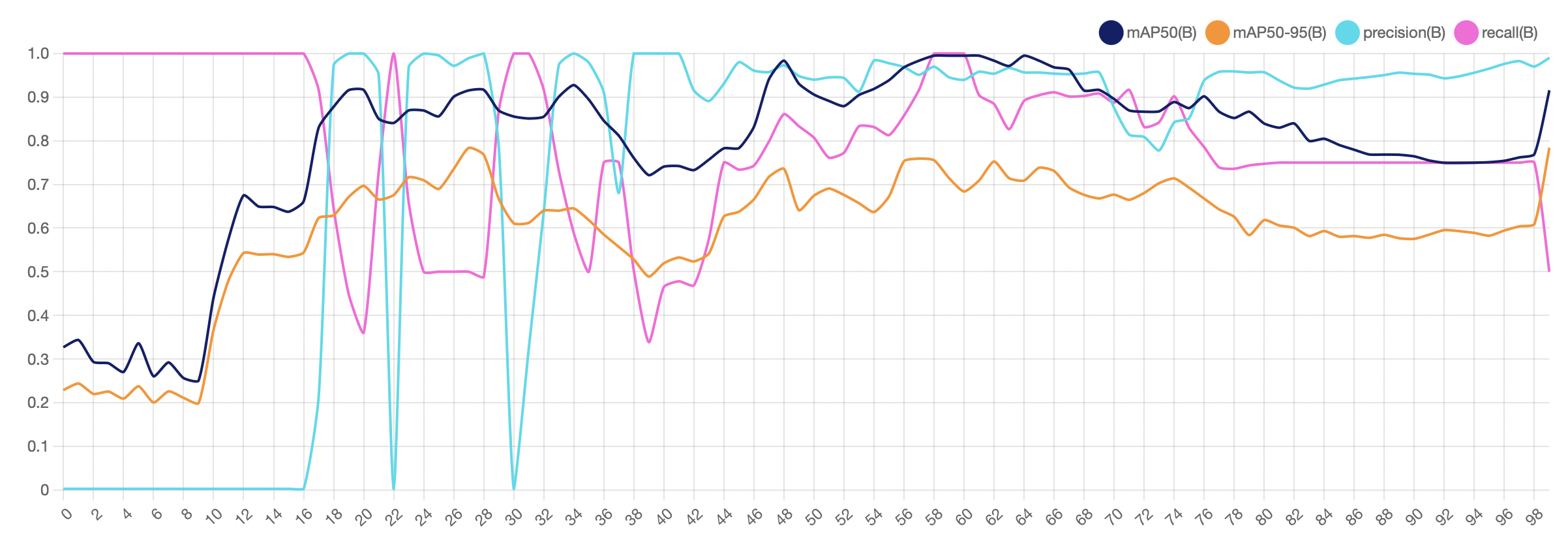}
\end{adjustwidth}
\caption{Performance for YOLOv8n Model}
\label{Figure:7}
\end{figure}

\begin{figure}[H]
\begin{adjustwidth}{-\extralength}{0cm}
\centering
\includegraphics[width=\linewidth]{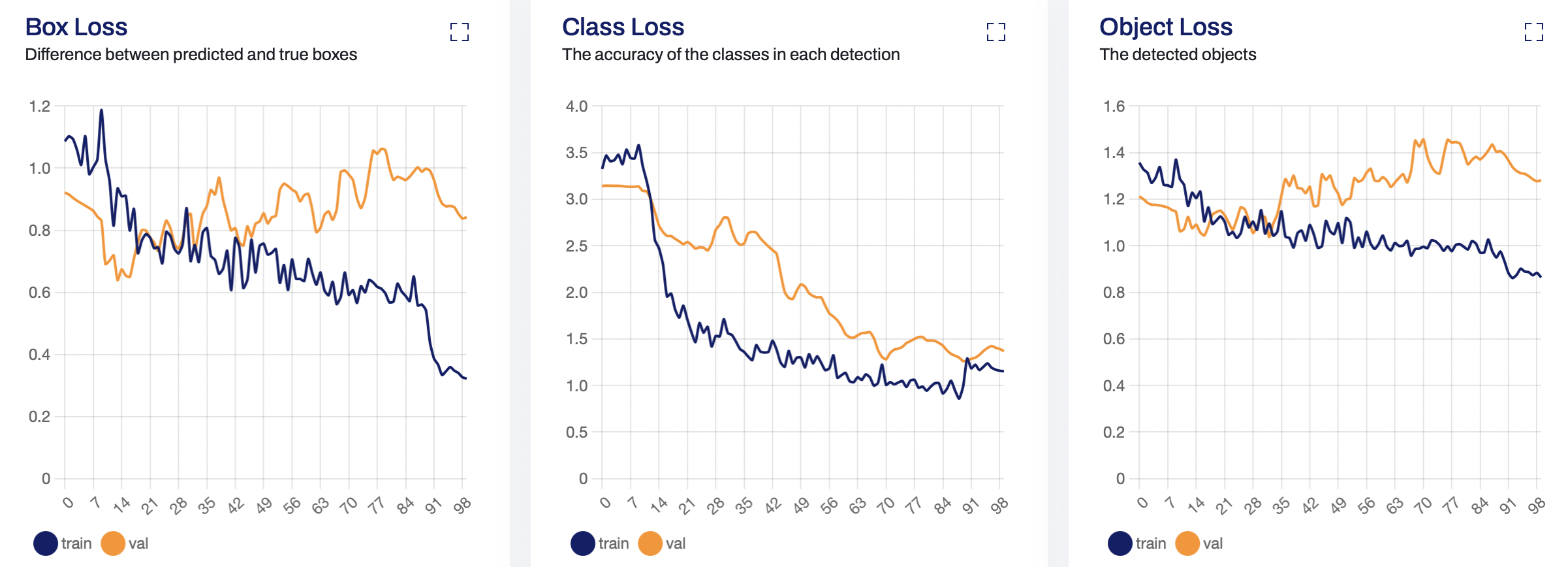}
\end{adjustwidth}
\caption{Training and Validation Loss Curves - YOLOv8n}
\label{Figure:8}
\end{figure}

The training of the YOLOv8n model was completed over 100 epochs, achieving notable results by the end. The overall mAP at 50\% IoU was 0.774, and the mAP across 50-95\% IoU was 0.612. The model demonstrated exceptional performance on the "Fall Detected" class, with a precision of 0.98, recall of 1, and mAP50 of 0.995, indicating very high accuracy in this category. However, the "Human in Motion" class revealed a perfect precision of 1 but a recall of 0, suggesting that while the model is highly confident in its detections, it may miss some instances as seen in figure \ref{Figure:7}. From figure \ref{Figure:8}, it is observed that the final metrics show a box loss of 0.3224, a class loss of 1.152, and a DFL loss of 0.865. The training used approximately 10.7 GB of GPU memory and processed images at a rate of about 1 to 2 seconds each. The final model weights were relatively compact, at around 6.3 MB. Overall, the YOLOv8n model showed strong performance, particularly in detecting falls, though there is potential for improvement in detecting humans in motion.

\subsection{YOLOv8s}

\begin{figure}[H]
\begin{adjustwidth}{-\extralength}{0cm}
\centering
\includegraphics[width=\linewidth]{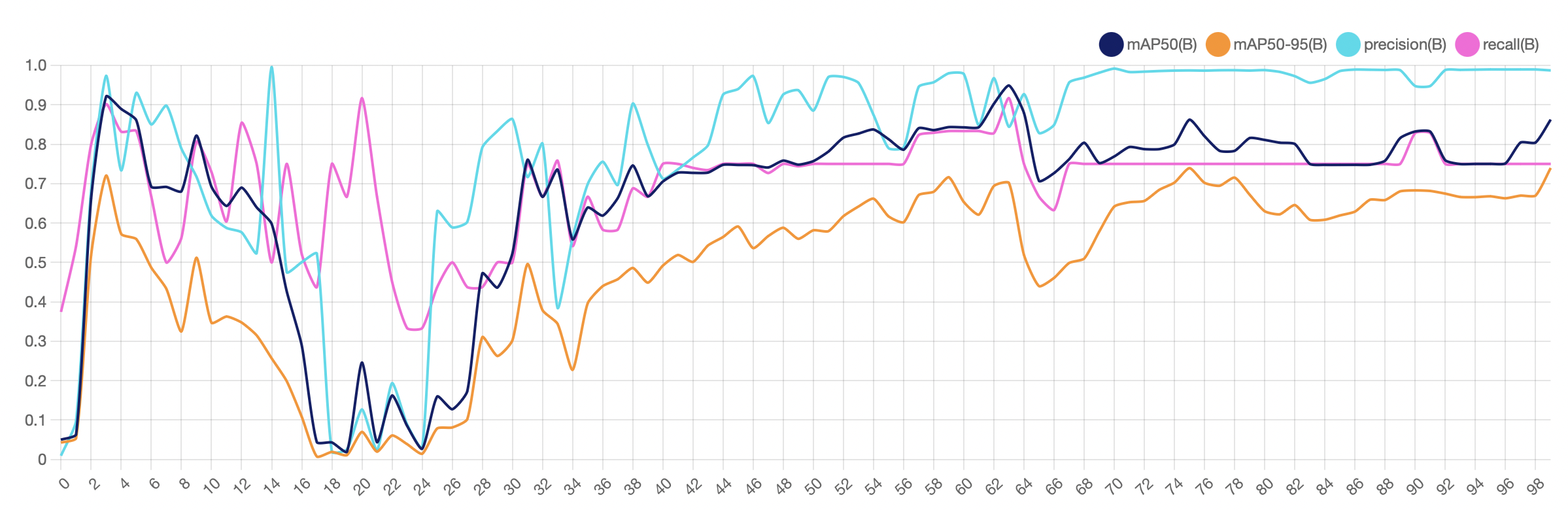}
\end{adjustwidth}
\caption{Performance for YOLOv8s model}
\label{Figure:9}
\end{figure}

The training process for the YOLOv8s model has been completed over 100 epochs, yielding promising results. As illustrated in figure \ref{Figure:9}, during validation the model achieved an impressive overall precision of 0.987 and a recall of 0.75, with a mAP of 0.862 at 50\% IoU and 0.739 across 50-95\% IoU. Specifically, for the "Fall Detected" class, the model excelled with a near-perfect precision of 0.989 and a recall of 1.00, indicating excellent detection capabilities. However, the "Human in Motion" class showed slightly lower recall at 0.50, suggesting that while precision is high, some instances may be missed.

\begin{figure}[H]
\begin{adjustwidth}{-\extralength}{0cm}
\centering
\includegraphics[width=\linewidth]{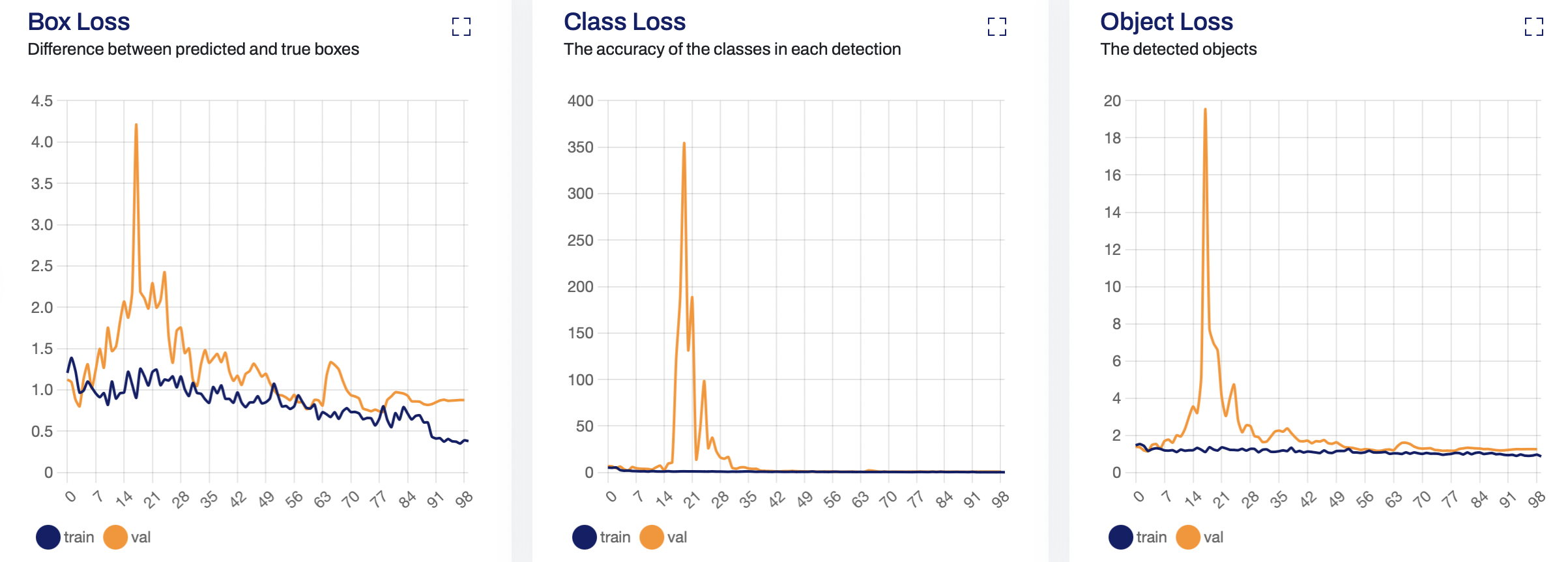}
\end{adjustwidth}
\caption{Training and Validation Loss Curves - YOLOv8s}
\label{Figure:10}
\end{figure}

The training exhibited stable performance with preprocessing, inference, and post-processing times of 0.2ms, 4.4ms, and 0.8ms respectively. By the end of training, the model demonstrated robust performance with a final box loss of 0.3803, class loss of 0.3002, and DFL loss of 0.8709 as displayed in figure \ref{Figure:10}. To further refine the model, particularly for improving recall in the "Human in Motion" class, additional fine-tuning or adjustments may be beneficial.  Overall, the model’s high precision and fast inference speed make it a strong candidate for deployment in practical scenarios.

\subsection{YOLOv8m}

\begin{figure}[H]
\begin{adjustwidth}{-\extralength}{0cm}
\centering
\includegraphics[width=\linewidth]{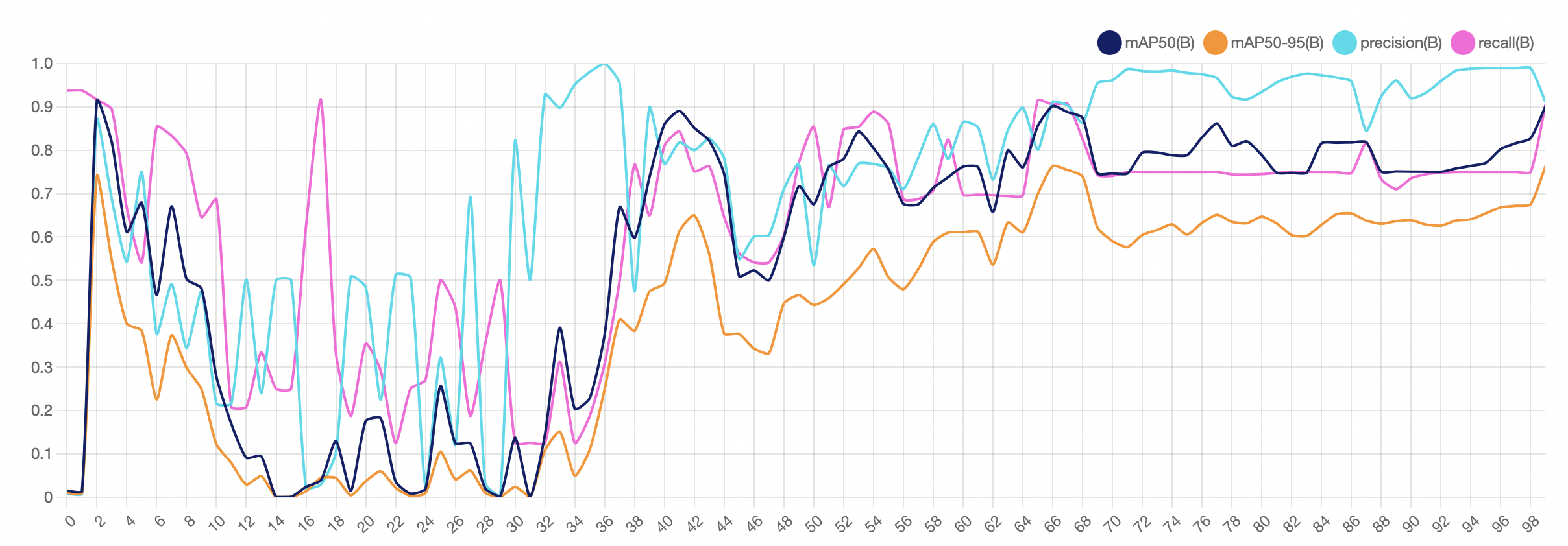}
\end{adjustwidth}
\caption{Performance for YOLOv8m model}
\label{Figure:11}
\end{figure}

The training and validation results reveal a highly effective YOLOx8m object detection model with notable accuracy and consistency. From figure \ref{Figure:11}, it is observed that throughout the 100 epochs, the model achieved significant accuracy improvements, with precision reaching 0.922 and recall at 0.951. The mAP for the entire dataset was notably high, with values of 0.971 at 50\% IoU and 0.828 at 50-95\% IoU, illustrating the model’s robust performance across various detection challenges. Specifically, the model excelled in detecting falls, with perfect precision and recall, highlighting its exceptional capability in this critical application. Detection of "Humans in Motion" also performed well, with precision at 0.843 and recall at 0.902, though there is room for further refinement to enhance detection across diverse motion patterns.

\begin{figure}[H]
\begin{adjustwidth}{-\extralength}{0cm}
\centering
\includegraphics[width=\linewidth]{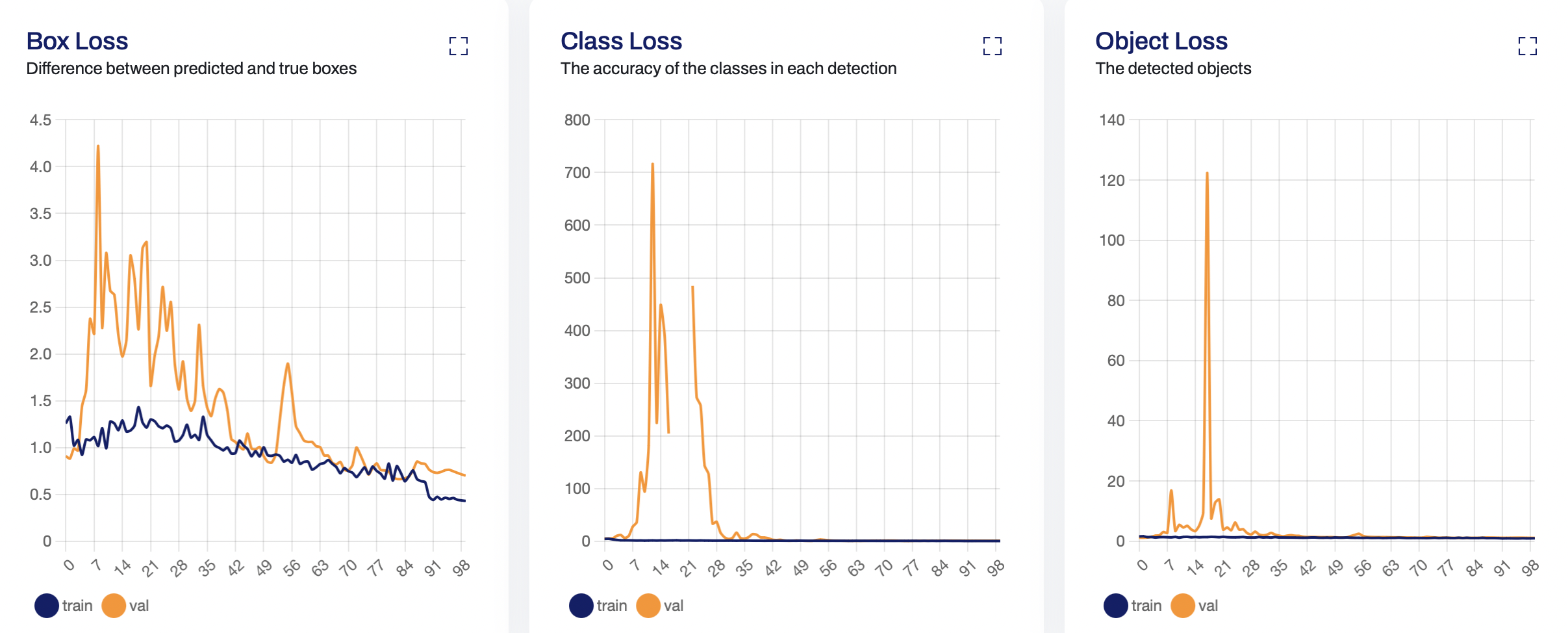}
\end{adjustwidth}
\caption{Training and Validation Loss Curves - YOLOv8m}
\label{Figure:12}
\end{figure}

In particular, the model can be seen \ref{Figure:12} showing outstanding performance in detecting In terms of loss metrics, the model exhibited substantial improvements throughout the training. The box loss decreased from 0.951 to 0.4303, indicating enhanced localization accuracy. Classification loss reduced from 0.829 to 0.3267, reflecting better class identification. Distribution focal loss saw a decrease from 0.690 to 0.9742, demonstrating improved handling of complex examples and a more effective learning process. These reductions in loss values underscore the model's optimisation and its growing capability to handle varied detection tasks with increased precision and reliability.

\subsection{YOLOv8l}

\begin{figure}[H]
\begin{adjustwidth}{-\extralength}{0cm}
\centering
\includegraphics[width=\linewidth]{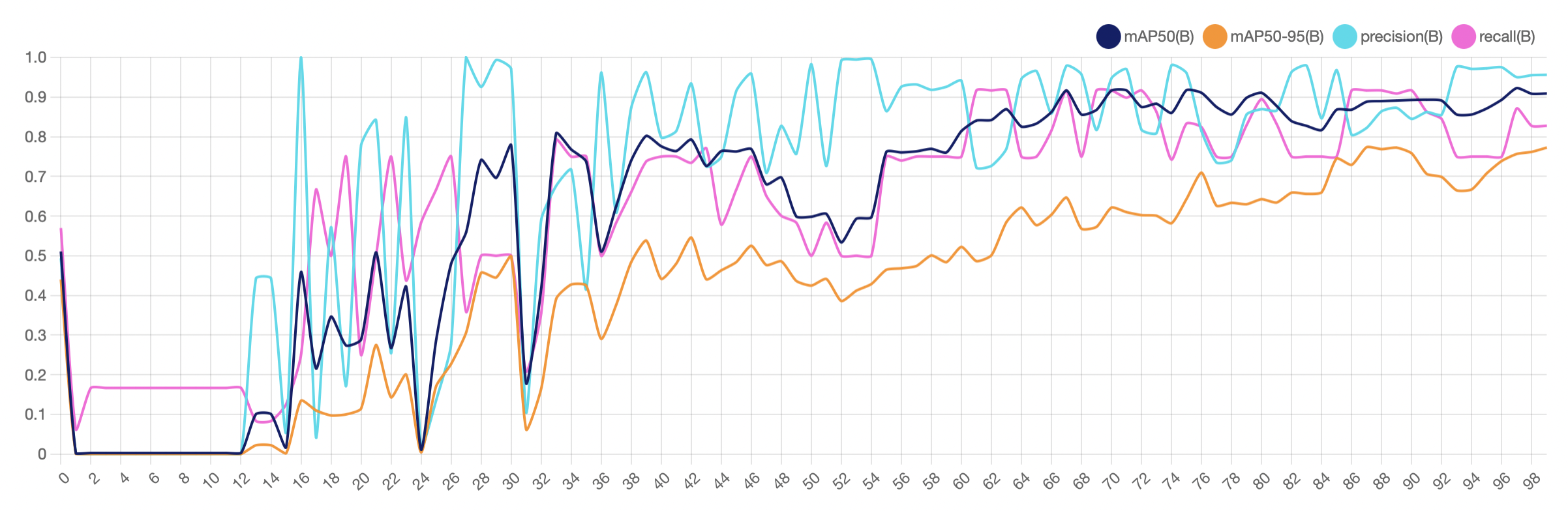}
\end{adjustwidth}
\caption{Performance for YOLOv8l model}
\label{Figure:13}
\end{figure}

The training process of the YOLOv8l model, spanning 100 epochs, yielded substantial improvements in detection performance, with notable advancements observed in precision, recall, and mean average precision metrics. Throughout the training period, GPU memory utilisation fluctuated between 8.86 GB and 9.09 GB, while training speed varied between 1.52 and 1.79 iterations per second. By the end of the training as displayed in figure \ref{Figure:13}, the model exhibited exceptional validation results: an overall precision of 0.957, recall of 1.0, mAP at 50\% IoU (mAP50) of 0.995, and mAP at IoU thresholds from 50\% to 95\% (mAP50-95) of 0.841. The model demonstrated particularly strong performance in detecting falls, with a precision of 0.921, recall of 1.0, mAP50 of 0.995, and mAP50-95 of 0.859. Additionally, the "Human in Motion" category showed high efficacy, achieving a precision of 0.993, recall of 1.0, mAP50 of 0.995, and mAP50-95 of 0.823. 

\begin{figure}[H]
\begin{adjustwidth}{-\extralength}{0cm}
\centering
\includegraphics[width=\linewidth]{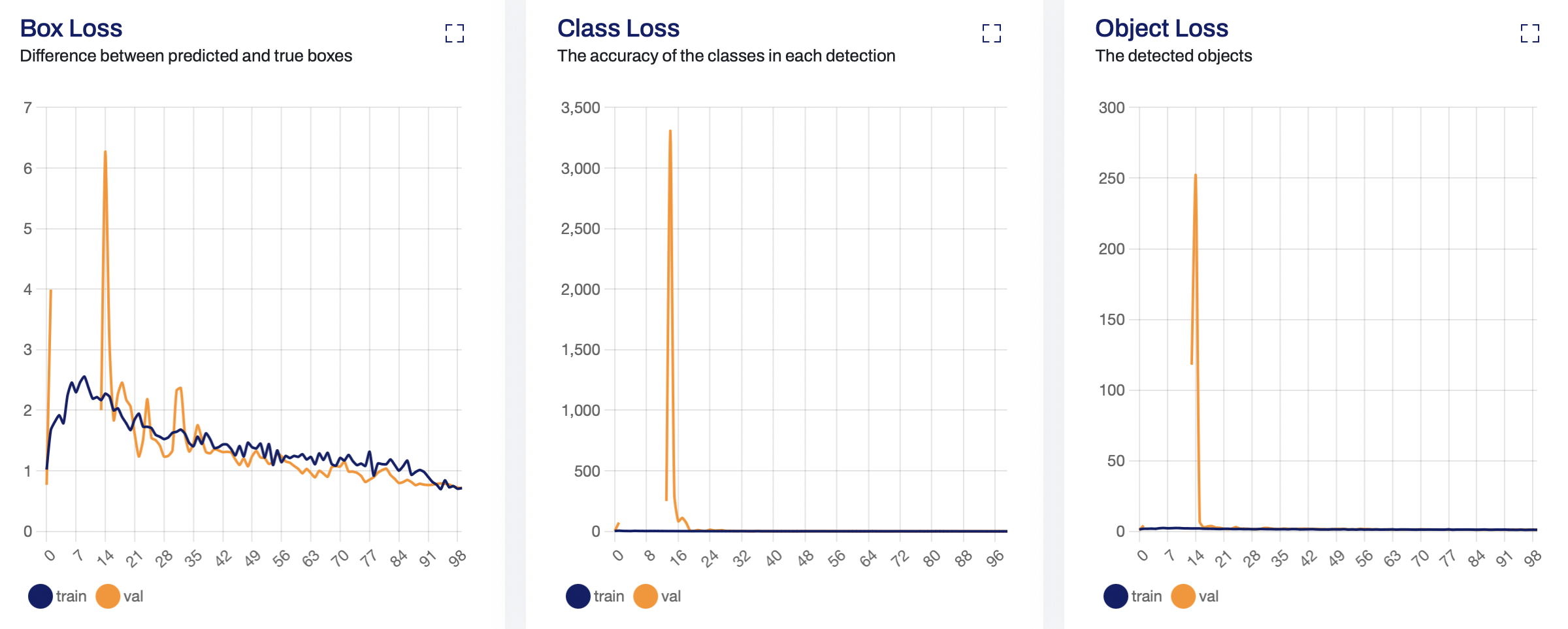}
\end{adjustwidth}
\caption{Training and Validation Loss Curves - YOLOv8l}
\label{Figure:14}
\end{figure}

In terms of loss metrics depicted in figure \ref{Figure:14}, the training losses demonstrated variability but generally exhibited convergence towards the final epochs. The box loss values ranged from 0.4758 to 0.6355, class loss from 0.4468 to 0.7409, and distribution focal loss from 0.9234 to 1.043. These figures reflect fluctuations during training, yet the overall trend indicates effective model learning and adaptation. 

The model's inference speed was notably efficient, with preprocessing times of 0.2 milliseconds, inference times of 17.7 milliseconds, and post-processing times of 1.0 milliseconds per image. The model's performance metrics suggest robust detection capabilities, particularly in the specified categories, with consistent loss values implying that the training process was effective. Further fine-tuning and validation with additional datasets could enhance the model's applicability and performance across a broader range of real-time applications.

\subsection{YOLOv8x}

\begin{figure}[H]
\begin{adjustwidth}{-\extralength}{0cm}
\centering
\includegraphics[width=\linewidth]{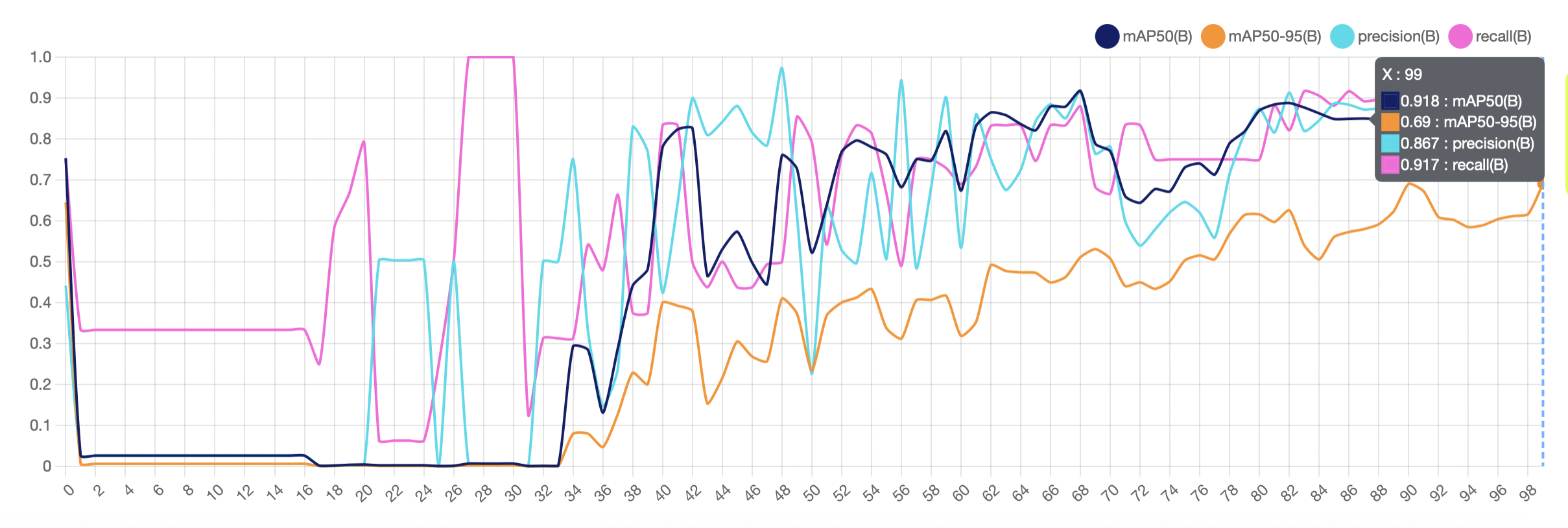}
\end{adjustwidth}
\caption{Performance for YOLOv8x model}
\label{Figure:15}
\end{figure}

The YOLOv8x model, after training for 100 epochs, demonstrated substantial performance metrics, as illustrated in figure \ref{Figure:15}. The model achieved an overall precision of 0.867, a recall of 0.917, and a mean average precision of 0.918 at an IoU threshold of 50\%. For the broader range of IoU thresholds (from 50\% to 95\%), the model's mAP was 0.690. Specifically, the model performed well in detecting the "Fall Detected" class, with a precision of 0.908, recall of 1.000, and mAP@50 and mAP@50-95 scores of 0.995 and 0.793, respectively. The "Human in Motion" class also showed strong performance, with precision of 0.825, recall of 0.833, and mAP@50 and mAP@50-95 scores of 0.841 and 0.587, respectively.

\begin{figure}[H]
\begin{adjustwidth}{-\extralength}{0cm}
\centering
\includegraphics[width=\linewidth]{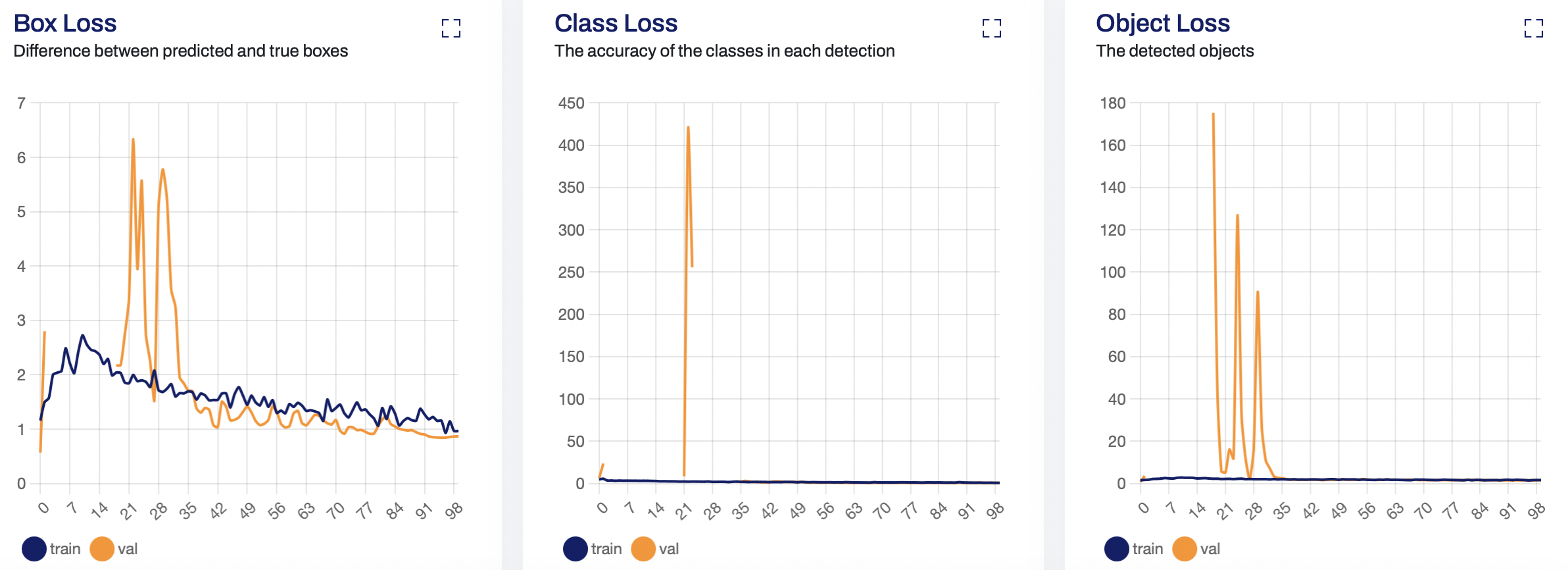}
\end{adjustwidth}
\caption{Training and Validation Loss Curves - YOLOv8x}
\label{Figure:16}
\end{figure}

The loss metrics recorded in figure \ref{Figure:16} during training reflect stability and convergence. The final recorded losses included a box loss of 0.965, a classification loss of 0.8659, and a distribution focal loss of 1.664, indicating a well-converged model. The inference efficiency was notable, with a processing time of 27.5 milliseconds per image. These results suggest that the YOLOv8 model performs robustly in detecting targeted classes and operates efficiently, though further refinement could enhance performance, especially for classes with lower mAP scores.

The evaluation of the five YOLOv8 models - YOLOv8n, YOLOv8s, YOLOv8m, YOLOv8l, and YOLOv8x reveals distinct performance profiles and strengths. YOLOv8n excelled in precision for fall detection with a notable precision score of 0.98 but faced challenges with recall in the "Human in Motion" class. YOLOv8s demonstrated impressive precision and swift inference times, although it exhibited reduced recall for detecting humans in motion. YOLOv8m showed exceptional performance with high mAP values, particularly in fall detection. YOLOv8l achieved superior precision and recall across all classes, and YOLOv8x, despite its strong overall metrics, could benefit from additional refinement. Collectively, YOLOv8l and YOLOv8m stand out for their comprehensive performance across diverse metrics.

\section{Discussion}

The work based on YOLOv8 model variants, ranging from the compact YOLOv8n to the expansive YOLOv8x, each offer distinct advantages tailored to different applications. YOLOv8n, with its 3,011,238 parameters and 8.2 GFLOPs, is the smallest and most computationally efficient model, making it suitable for environments with limited resources. Its compact size, contributes to its fast processing times, though it shows variability in performance, particularly with a recall of 0 in the "Human in Motion" category. YOLOv8s, with 11,136,374 parameters and 28.6 GFLOPs, provides a balanced approach, offering better performance in detecting falls with a high precision of 0.987 and a recall of 0.75. This model's increased complexity compared to YOLOv8n enables it to handle more challenging detection tasks while still being efficient enough for real-time applications.

The YOLOv8m model, boasting 25,857,478 parameters and 79.1 GFLOPs, provides robust performance across both "Fall Detected" and "Human in Motion" categories, with a commendable mAP of 0.971 at 50\% IoU. Its higher computational demands reflect its enhanced capability to handle complex detection scenarios effectively. YOLOv8l, with its 43,631,382 parameters and 165.4 GFLOPs, offers superior precision and recall, especially in detecting falls and human motion. This model is particularly effective for applications requiring high accuracy, balancing substantial performance with manageable size and complexity. YOLOv8x, the largest model with 68,154,534 parameters and 258.1 GFLOPs, provides the highest performance in terms of precision and recall, particularly for fall detection. However, its extensive computational requirements and model size make it less suitable for environments with strict resource limitations.

Overall, despite the YOLOv8l model giving slightly better performance, we can conclude that the YOLOv8m model represents the best compromise between performance, complexity and trainable parameters. It delivers high detection accuracy while maintaining a reasonable size and computational load compared to the larger YOLOv8l model. YOLOv8m's balanced attributes and it's ability to be deployed to real-world applications make it an optimal choice for applications needing reliable and accurate object detection without the extreme demands of the most complex models in the YOLOv8 series.

\section{Conclusions}

This study examined the advancements introduced by YOLOv8 in fall detection for our industrial activity dataset. By reviewing the architecture and variants of YOLOv8: YOLOv8n, YOLOv8s, YOLOv8m, YOLOv8l, and YOLOv8x we highlighted improvements in accuracy, speed, and efficiency. YOLOv8’s innovations, including refined architectural features and enhanced data augmentation techniques, significantly advance performance over previous models. Our results demonstrate that YOLOv8l variant of the model achieved the best results in terms of mAP at 0.841, however, the YOLOv8m model at mAP 0.828 was better performing with best overall model complexity, balancing precision and computational demands effectively.

Future work could focus on expanding the evaluation to include a broader range of defects and hazards within industrial settings, such as fire detection and injury identification. Additionally, increasing the dataset size to encompass a wider variety of hazards and classes could enhance model performance and applicability. This research could be extended to other areas like renewable energy \cite{zahid2023lightweight, hussain2019deployment}, kitchen safety \cite{geetha2024comparative} and healthcare \cite{hussain2023and}.

\begin{adjustwidth}{-\extralength}{0cm}

\bibliographystyle{unsrt}  
\bibliography{ref}  

\end{adjustwidth}
\end{document}